\documentclass{article} % For LaTeX2e
\usepackage{iclr2021_conference,times}
\usepackage{float}
\usepackage{graphicx}
\usepackage{amsmath,amsfonts,bm}

\def\eqref#1{equation~\ref{#1}}
\def\1{\bm{1}}

\DeclareMathAlphabet{\mathsfit}{\encodingdefault}{\sfdefault}{m}{sl}
\SetMathAlphabet{\mathsfit}{bold}{\encodingdefault}{\sfdefault}{bx}{n}

\usepackage{hyperref}
\usepackage{url}

\usepackage{xspace}
\def\onedot{.\xspace}
\def\eg{\emph{e.g}\onedot} 

\def\ie{\emph{i.e}\onedot}

\title{
Rethinking Hierarchies in Pre-trained Plain Vision Transformer
}

\author{
Yufei Xu$^{1}$, \quad
Jing Zhang$^{1}$, \quad
Qiming Zhang$^{1}$, \quad
Dacheng Tao$^{2,1}$ \quad
\vspace{1 mm}
\\
\vspace{1 mm}
\textsuperscript{1}School of Computer Science, The University of Sydney, Australia \\
\textsuperscript{2}JD Explore Academy, China \\
}

\iclrfinalcopy % Uncomment for camera-ready version, but NOT for submission.
\begin{document}

\maketitle

\begin{abstract}

Self-supervised pre-training vision transformer (ViT) via masked image modeling (MIM) has been proven very effective. However, customized algorithms should be carefully designed for the hierarchical ViTs, e.g., GreenMIM, instead of using the vanilla and simple MAE for the plain ViT. More importantly, since these hierarchical ViTs cannot reuse the off-the-shelf pre-trained weights of the plain ViTs, the requirement of pre-training them leads to a massive amount of computational cost, thereby incurring both algorithmic and computational complexity. In this paper, we address this problem by proposing a novel idea of disentangling the hierarchical architecture design from the self-supervised pre-training. We transform the plain ViT into a hierarchical one with minimal changes. Technically, we change the stride of linear embedding layer from 16 to 4 and add convolution (or simple average) pooling layers between the transformer blocks, thereby reducing the feature size from 1/4 to 1/32 sequentially. Despite its simplicity, it outperforms the plain ViT baseline in classification, detection, and segmentation tasks on ImageNet, MS COCO, Cityscapes, and ADE20K benchmarks, respectively. We hope this preliminary study could draw more attention from the community on developing effective (hierarchical) ViTs while avoiding the pre-training cost by leveraging the off-the-shelf checkpoints. The code and models will be released at \href{https://github.com/ViTAE-Transformer/HPViT}{https://github.com/ViTAE-Transformer/HPViT}.

\end{abstract}

\section{Introduction}

By training the networks to reconstruct the masked input, masked image modeling (MIM)~\cite{bao2022beit,MaskedAutoencoders2021,xie2022simmim} has proven to be very effective for pre-training vision transformers~\cite{dosovitskiy2020image}. The representative method, \eg, MAE~\cite{MaskedAutoencoders2021}, improves the efficiency of the pre-training process by only operating on the visible tokens. However, such a design breaks the 2D relationship between tokens, making it nontrivial to directly pre-training hierarchical vision transformers.

To leverage MIM for pre-training hierarchical vision transformers, recent studies propose customized algorithms like token grouping~\cite{huang2022green}, uniform masking strategy~\cite{li2022uniform}, and mixed masking strategy with masked attention~\cite{liu2022mixmim}. Some works also introduce specific layers~\cite{zhang2022hivit} to facilitate the learning of hierarchical vision transformers in MIM pre-training. {Although} these methods achieve excellent performance, they {need elaborate designs compared to MAE} and inevitably require training the networks from scratch {rather than making good use of off-the-shelf plain vision transformer checkpoints, due to the} discrepancy between the hierarchical and plain architectures, e.g., spatial down-sampling. Consequently, the pre-training process introduces a massive amount of computational cost {and incurs} both algorithmic and computational complexity. 
Such a difficulty hinders the extensive study of the more effective and efficient designs of hierarchical vision transformers, especially the design of larger models. 

To address this problem, we propose a novel idea of disentangling the hierarchical architecture design from the MIM pre-training process. Specifically, we directly transform the MIM pre-trained plain vision transformers into hierarchical ones with minimal changes, thereby alleviating the algorithmic and computational complexity of pre-training the hierarchical vision transformers from scratch. Technically, we leverage checkpoints provided by MAE~\cite{MaskedAutoencoders2021} in the experiments. We {reduce} the stride of the linear embedding layer from 16 to 4 and add convolution {or} pooling layers between the transformer blocks, thereby reducing the feature size from 1/4 to 1/32 {gradually} and obtaining the hierarchical structures.
It should be noted that thanks to the flexibility of plain vision transformers, researchers can easily determine the transformed hierarchical structure, \eg, by controlling the number of layers in each stage, to explore effective and efficient configurations of hierarchical vision transformers, without the need of re-training.
Compared with the highly entangled baseline, \ie, using plain vision transformers in both pre-training and fine-tuning, our transformed hierarchical transformers obtain better performance in various vision tasks, including classification on ImageNet~\cite{deng2009imagenet}, detection on MS COCO~\cite{lin2014microsoft}, and segmentation on Cityscapes~\cite{cordts2016cityscapes} and ADE20K~\cite{ade20k}, {without the} cost {of} pre-training from scratch. It demonstrates the benefits of disentangling the architecture design from the MIM pre-training and we hope it will inspire more research in this direction.

\section{Methods}
\begin{figure}[t]
    \centering
    \includegraphics[width=\linewidth]{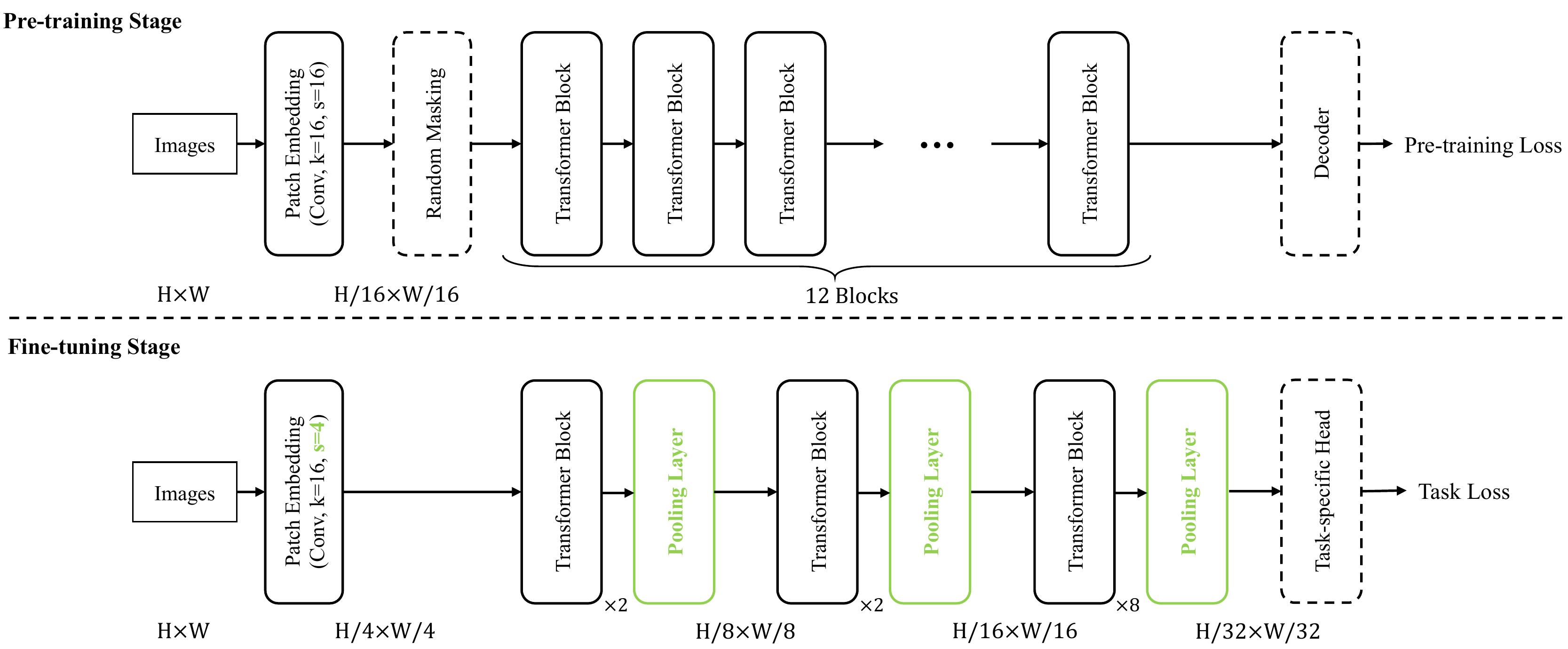}
    \caption{Illustration of the proposed method given pre-trained ViT-B. The dashed boxes indicate the operations are only used in the corresponding stage. The black boxes represents the blocks whose parameters are reused during fine-tuning. We use green to mark the changes that we introduce to make the plain vision transformers hierarchical.}
    \label{fig:frameWork}
\end{figure}

In this paper, we focus on disentangling the structures between pre-training and fine-tuning and designing hierarchical vision transformers given pre-trained plain vision transformers. As demonstrated in Figure~\ref{fig:frameWork}, we only transform the structure of the vision transformer in the fine-tuning stages through some simple yet flexible designs. 
{All parts of the encoder in} the pre-trained plain vision transformers are reused in the {transformed} hierarchical architectures {and only minimal changes are introduced} (marked by green). We will {give these changes in detail in the following parts}.

\textbf{Patch Embedding Layer} To embed the input images, plain vision transformers partition the inputs into non-overlapping patches first and then project them to tokens~\cite{dosovitskiy2020image}. The typical size of each patch is 16, i.e., the feature map is spatially down-sampled by 16$\times$. In this case, some high-frequency information is {discarded}. To recover the high-resolution features and reuse the parameters, we partition the input image into overlapped patches while retaining the size of each patch, \ie, we partition the image into 16$\times$16 patches and allow adjacent patches to have an overlap of size 12, {resulting in 4$\times$ down-sampling in the patch embedding layer.} The overlapped patches are then projected to tokens using the weights from the pre-trained embedding layer directly.

\textbf{Window-based Attention} Directly increasing the size of the feature map from 1/16 to 1/4 will introduce {extremely} heavy computational burden due to the quadratic computational complexity of self-attention. To address this issue, we change the attention calculation from the vanilla (global) self-attention to the (local) window-based one while making no changes to the projection weights of query, key, and value. Specifically, we follow the common practice in recent studies~\cite{liu2021swin,zhang2022vitaev2,li2022exploring,xu2022vitpose} to adopt window-based attention with relative position embedding. It should be noted {that} we only adopt the original window-based attention as a proof of concept while there are also several advanced attention manners that could be used to further improve the performance, \eg, VSA~\cite{zhang2022vsa} and {grid} attention~\cite{tu2022maxvit}. We leave it as future work.

\textbf{Pooling layer} One important design in hierarchical vision transformer is that we need to down-sample the feature maps {gradually} to obtain hierarchical feature representations. We start from the parameter-free average pooling operations to down-sample the feature maps. To help broadcast the information between adjacent patches, we use slightly larger kernels in the pooling layer, \ie, using $3 \times 3$ kernel size and stride 2. A parallel convolution layer with the same kernel size is introduced to help learn better local features. The parallel convolution and pooling layers can be merged after training using the re-parameterization trick~\cite{ding2021repvgg} {for inference speedup}.

\textbf{Discussion} This work aims to transform the structure of the plain vision transformers during the fine-tuning stage by exploring the flexibility of the transformer architecture as well as considering the properties of downstream computer vision tasks. It focuses on a different aspect compared with the efficient fine-tuning methods originally proposed in the NLP area, \eg, using adapters~\cite{vitadaper} to adapt the transformers' outputs or prompts~\cite{jia2022visual} to tune the networks' inputs. These methods are orthogonal to ours and can be used to further improve the performance.

\section{Experiments}
To thoroughly validate the effectiveness of the proposed method, we evaluate the performance of the transformed hierarchical vision transformer on classification, detection, and segmentation tasks. We use the pre-trained ViT-B~\cite{dosovitskiy2020image} checkpoints from MAE~\cite{MaskedAutoencoders2021} as our baseline and transform the backbone network into a hierarchical one during fine-tuning. We use window attention with a window size of $7 \times 7$ in the 1/4 and 1/8 levels, respectively.

\begin{table}[htbp]
  \small
  \centering
  \caption{Classification results. $^\dag$ represents we use transformed hierarchical structures based on the pre-trained ViT-B. We adopt GreenMIM~\cite{huang2022green}, HiViT~\cite{zhang2022hivit}, ViTAE~\cite{xu2021vitae}, BEiT~\cite{bao2022beit}, and MAE~\cite{MaskedAutoencoders2021} for comparison.}
   {\begin{tabular}{c|cccccc}
    \hline
    Method & GreenMIM & HiViT & ViTAE & BEiT  & MAE   & Ours \\
    \hline
    Pre-train  & Swin-B & HiViT & ViTAE-B & ViT-B & ViT-B & ViT-B \\
    Fine-tune & Swin-B & HiViT & ViTAE-B & ViT-B & ViT-B & ViT-B$^\dag$ \\
    \hline
    Accuracy & 83.8  & 83.8  & 83.8 & 83.2  & 83.6  & 83.8 \\
    \hline
    \end{tabular}}%
  \label{tab:hvit_classification}%
\end{table}%

\textbf{Classification} We fine-tune the transformed model for 100 epochs on the ImageNet~\cite{deng2009imagenet} dataset. The last pooling layer of the transformed model is modified to a simple global average pooling. The results are presented in Table~\ref{tab:hvit_classification}. It can be observed that simply transforming the vision transformer from a plain structure to a hierarchical one brings about 0.2 performance gains in accuracy. Such performance is comparable with the performance using the hierarchical structure-friendly pre-training, \eg, {83.8 of} Swin-B~\cite{liu2021swin} pre-trained with GreenMIM~\cite{huang2022green} and HiViT~\cite{zhang2022hivit}. Unlike them, our method does not require the structures to be the same during pre-training and fine-tuning, thus getting rid of the {pre-training} cost. 

\begin{table}[htbp]
  \centering
  \small
  \caption{Detection results based on the Mask RCNN~\cite{he2017mask} 1$\times$ setting on MS COCO~\cite{lin2014microsoft}.}
    \begin{tabular}{c|cccccc}
    \hline
          & $mAP^b$    & $mAP^b_{50}$  & $mAP^b_{75}$  & $mAP^m$    & $mAP^m_{50}$    & $mAP^m_{75}$ \\
    \hline
    ViT-B & 44.0    & 66.2  & 48.2  & 40.7  & 63.4  & 43.8 \\
    ViT-B$^\dag$ & 45.5  & 67.1  & 50.2  & 41.5  & 64.4  & 44.9 \\
    \hline
    \end{tabular}%
  \label{tab:hvit_det}%
\end{table}%

\textbf{Detection} We use Mask RCNN~\cite{he2017mask} as the detection framework. {We use} the plain vision transformer and the transformed one as the backbone {for comparison}. The experiments are conducted on the MS COCO dataset~\cite{lin2014microsoft} and the models are trained for 12 epochs, following the default 1$\times$ setting \cite{liu2021swin}. We follow the strategy in \cite{li2021benchmarking} to generate hierarchical features from the plain vision transformer via transposed convolutions. Four global attention layers are used evenly in the baseline model. {To reduce the computational cost}, we {replace} the first global attention layer in the transformed hierarchical backbone {with window attention}. %since its location is within the high-resolution stages in the transformed architecture. 
The last three global attention layers are retained at the same position as the baseline model. The output features are generated from the 2nd, 4th, 10th, and 12th layers, following the common practice in hierarchical transformers~\cite{liu2021swin}. As observed from Table~\ref{tab:hvit_det}, the transformed backbone brings a gain of 1.5 box mAP and 0.8 mask mAP over the baseline model. The results demonstrate that the hierarchical structure is more suitable for dense prediction tasks compared with the plain vision transformer baseline. Besides, {the proposed method can make good use of the off-the-shelf checkpoints of pre-trained plain vision transformers, avoiding bringing the extra cost of pre-training the model from scratch.}
%   can provide a good initialization for hierarchical vision transformers without . 

\begin{table}[htbp]
  \centering
  \small
  \caption{Segmentation results based on the UperNet~\cite{xiao2018unified} on Cityscapes~\cite{cordts2016cityscapes} (769 $\times$ 769 input size and 40K training iterations) and ADE20K~\cite{ade20k} (640 $\times$ 640 input size and 80K training iterations).}
    \begin{tabular}{c|cc|cc}
    \hline
          & \multicolumn{2}{c|}{Cityscapes} & \multicolumn{2}{c}{ADE20K} \\
          & mIoU  & mAcc  & mIoU  & mAcc \\
    \hline
    ViT-B & 80.0    & 87.2  & 47.3  & 58.8 \\
    ViT-B$^\dag$ & 81.2  & 88.3  & 48.1  & 59.8 \\
    \hline
    \end{tabular}%
  \label{tab:hvit_seg}%
\end{table}%

\textbf{Segmentation} To evaluate the performance of the transformed vision transformer on the segmentation task, we adopt UperNet~\cite{xiao2018unified} as the segmentation framework and conduct experiments on the Cityscapes~\cite{cordts2016cityscapes} and ADE20K~\cite{ade20k} datasets, respectively. We adopt the strategy in BEiT~\cite{bao2022beit} to extract the output features from the 4th, 6th, 8th, and 12th layers of the plain vision transformer with transposed convolutions. As in the detection tasks, the output features are generated from the 2nd, 4th, 10th, and 12th layers from the transformed hierarchical vision transformer. We employ simple max pooling layers in the last pooling layer with kernel size 2 and stride 2. As shown in Table~\ref{tab:hvit_seg}, the transformed model improves the baseline model {by} 0.8 mIoU on the ADE20K dataset and 1.2 mIoU on the Cityscapes dataset. These results further confirm the value of the proposed simple idea, i.e., the hierarchical vision transformer transformed from the {pre-trained} plain vision transformer is more suitable for dense prediction tasks since {the hierarchical design keeps valuable high-resolution features in early stages and explicitly produces multi-scale features for the task heads.}
% takes more vision inductive bias into consideration. 

\section{Limitation and Discussion}

This study provides a promising direction that deserves more research efforts. More effective hierarchical structures are expected to be derived from the plain vision transformers, \eg, by trying more advanced attention or down-sampling methods, which should deliver better performance on general or specific downstream tasks. More importantly, they do not need re-training. For now, this work only focuses on transforming plain vision transformers pre-trained using self-supervised learning methods, especially based on masked image modeling. However, since there is no assumption about a specific pre-training method, it can be extended to transforming other pre-trained plain vision transformers, e.g., via cross-modal pre-training like CLIP~\cite{radford2021learning}. We leave it as future work.

\section{Conclusion}

In this paper, we rethink the masked image modeling pre-training for hierarchical vision transformers and present a novel idea of disentangling the hierarchical vision transformer design from pre-training. With simple pooling layers, the transformed hierarchical backbone outperforms the plain baseline in classification, detection, and segmentation tasks. Besides, such a method can fully utilize the progress in plain vision transformer pre-training for better hierarchical vision transformer fine-tuning with no requirements of expensive {pre-training}, contributing to the development of green AI {and providing a new path for the vision transformer research using limited computational resources}. We hope this preliminary study could draw more attention from the community on developing more effective vision transformers while avoiding the {pre-training} cost by exploiting the amazing flexibility of the transformer architecture.

\footnotesize{
\bibliographystyle{iclr2021_conference}
\bibliography{egbib}
}

\appendix

\end{document}